\documentclass[conference]{IEEEtran}
\IEEEoverridecommandlockouts
\usepackage{cite}
\usepackage{amsmath,amssymb,amsfonts}
\usepackage{algorithmic}
\usepackage{graphicx}
\usepackage{textcomp}
\usepackage{xcolor}
\usepackage{hyperref}
\usepackage{xspace}
\usepackage{amsmath}
\usepackage{placeins}
\usepackage{standalone}
\usepackage{tabularx}
\usepackage{mathtools}
\usepackage{tikz}
\usetikzlibrary{decorations.pathreplacing}
\usepackage{float}
\usepackage{booktabs}
\usepackage[light]{antpolt}
\usepackage[T1]{fontenc}
\usepackage{ulem}
\newcommand{\upperRomannumeral}[1]{\uppercase\expandafter{\romannumeral#1}}

\hypersetup{
    colorlinks=true,
    linkcolor=blue,
    filecolor=magenta,      
    urlcolor=cyan,
    pdftitle={},
    pdfpagemode=FullScreen,
    }

\newcommand{\comment}[1]{} 
\begin{document}

\title{CNN-Trans-Enc: A CNN-Enhanced Transformer-Encoder On Top Of Static BERT representations for Document Classification} 

\author{\IEEEauthorblockN{Charaf Eddine Benarab}
\IEEEauthorblockA{\textit{School of Computer Science and Engineering}\\
\textit{University of Electronics Science and Technology of China} \\\
Chengdu, China \\
charafeddineben@std.uestc.edu.cn}
\and
\IEEEauthorblockN{Shenglin Gui}
\IEEEauthorblockA{\textit{School of Computer Science and Engineering}\\
\textit{University of Electronics Science and Technology of China} \\\
Chengdu, China \\
shenglin\_gui@uestc.edu.cn}
}

\maketitle
\begin{abstract}
BERT achieves remarkable results in text classification tasks, it is yet not fully exploited, since only the last layer is used as a representation output for downstream classifiers. The most recent studies on the nature of linguistic features learned by BERT, suggest that different layers focus on different kinds of linguistic features. We propose a CNN-Enhanced Transformer-Encoder model which is trained on top of fixed BERT $[CLS]$ representations from all layers, employing Convolutional Neural Networks to generate QKV feature maps inside the Transformer-Encoder, instead of linear projections of the input into the embedding space. CNN-Trans-Enc is relatively small as a downstream classifier and doesn't require any fine-tuning of BERT, as it ensures an optimal use of the $[CLS]$ representations from all layers, leveraging different linguistic features with more meaningful, and generalizable QKV representations of the input. Using BERT with CNN-Trans-Enc keeps $98.9\%$ and $94.8\%$ of current state-of-the-art performance on the IMDB and SST-5 datasets respectably, while obtaining new state-of-the-art on YELP-5 with $82.23$ ($8.9\%$ improvement), and on Amazon-Polarity with $0.98\%$ ($0.2\%$ improvement) (K-fold Cross Validation on a 1M sample subset from both datasets). On the AG news dataset CNN-Trans-Enc achieves $99.94\%$ of the current state-of-the-art, and achieves a new top performance with an average accuracy of $99.51\%$ on DBPedia-14.
\end{abstract}

\section{Introduction}
Transformer based architectures\cite{attention}, and BERT\cite{Bert} in particular achieve remarkable results in natural language processing. Within document classification, and although BERT achieves outstanding results on almost all datasets available, there is still potential to further explore the effect that the classifier added on top of the last layer has on the performance. BERT contains multiple Transformer-Encoder layers~\cite{attention}, which output different hidden states for the same input. The experiment conducted by Jawahar et al.~\cite{jawahar-etal-2019-bert}, provides an empirical diagnosis of the linguistic structures learned by different BERT layers. Their experiment employs probing tasks~\cite{conneau-etal-2018-cram}, where specific auxiliary classifiers are used to predict targeted properties in a sentence representation. Taking the activation of a $[CLS]$ token, which is representative of the whole sentence due to the attention mechanism embedded in the Transformer-Encoder~\cite{attention}, and primarily used in classification tasks. 
Jawahar et al.~\cite{jawahar-etal-2019-bert} show that after probing BERT layers separately, it is observed that three different kinds of linguistic features are better encoded by different layers. Their findings which are prerequisites in this paper are as follows: Surface features are mostly captured by bottom layers, syntactic features by middle layers, and semantic features by higher layers. 

We propose a variant of the Transformer-Encoder layer introduced by Vaswani et al.~\cite{attention}, replacing Key, Query, and Value projection matrices with convolution-based submodules. The input to our CNN-Trans-Enc model consists of $[CLS]$ representations from all layers of BERT~\cite{Bert}, this ensures the downstream model is receiving multiple representations which are rich in different kinds of linguistic features, as suggested by Jawahar et al.~\cite{jawahar-etal-2019-bert}

We compare the proposed CNN-enhanced Transformer-Encoder to the CNN for Sentence Classification~\cite{kim2014convolutional}, vanilla Transformer-Encoder~\cite{attention}, and a simple Softmax classifier~\cite{howtofinetunbert} (Which is the common approach).
We train the Softmax classifier exclusively on the last $[CLS]$ token representation, the CNN-based models both on the last $[CLS]$ representation and on all $[CLS]$ representations from all layers, and the Transformer-Encoder and CNN-enhanced Transformer-Encoder on all $[CLS]$ representations obtained from all layers. The contributions of this paper are summarized as follows:

\begin{itemize}
    \item We demonstrate the importance of linguistic feature types in the task of Sentiment/Document Classification. We build on a conjecture that suggests using representations rich in different linguistic features would improve classification performance, and we investigate our assumption by training different models on fixed BERT representations learned by the last layer, and all layers ($[CLS]$ representations). 
    \item We introduce an approach which employs $[CLS]$ representations from different layers of BERT as channels to a CNN sub-architecture, congenially with a Transformer-Encoder for the purpose of leveraging all BERT-layers to generate an optimal representation for classification.
    \item BERT+CNN-Trans-Enc exceeds the state-of-the-art results on YELP-5, Amazon-Polarity, and DBPedia-14 datasets, and outperforms most models on all used datasets in our experiment without any kind of fine-tuning of BERT. 
    \end{itemize}

\section{Related work}

Text Classification~\cite{shengsurvey} in the context of deep learning mainly includes: Sentiment Analysis, News Categorization, and Topic Classification. To validate and test our method, we use six different datasets spanning the three tasks mainly constituting Text Classification. Survey~\cite{shengsurvey} reviews multiple datasets and models employed within the context of Text Classification (using deep learning models).

BERT (Bidirectional Encoder Representations from Transformers) ~\cite{Bert} is a landmark architecture, which is fundamentally a stack of bidirectional Transformer-Encoders~\cite{attention}, pre-trained for masked language and next sentence prediction. BERT learns sentence representations which are semantically rich, and achieves remarkable results on a variety of language-related benchmarks. BERT takes a sequence of tokens as input, and outputs a representation for the whole input after adding special tokens signifying the beginning and end of a sentence. For classification, the most common practice is to feed the $[CLS]$ representation obtained from the last layer (hidden-state of the $[CLS]$ token from the last Transformer-Encoder layer), into a Softmax~\cite{howtofinetunbert}, which outputs the probability of a label $c$:
\begin{equation}\label{eq:Softmax}
    p(c|h) = Softmax(\mathbf{W} \cdot h_{[CLS]_{12}})
\end{equation}
$\mathbf{W}$ is a task-specific parameter matrix, and $h_{[CLS]_{12}}$ is the last hidden-state of the $[CLS]$ token.\newline\indent
Pre-training and fine-tuning BERT on downstream datasets can be computationally expensive, given the large number of parameter it contains, and the ambiguity behind why it works so well. ALBERT~\cite{albert} proposes a lightweight version of BERT employing parameter-reduction techniques to lower memory consumption, this shows significant improvements in scaling on multi-sentence inputs. RoBERTa~\cite{liu2019roberta} on the other hand presents a study of BERT pre-training, and applies alternative pre-training objectives and tasks, showcasing the effect they have (alongside pre-training data) have on the performance. SMART~\cite{smart} proposes a learning framework to achieve efficient fine-tuning and attain better generalization, using smoothness-inducing regularization and Bregman proximal point optimization. SMART\textsubscript{RoBERTa}~\cite{smart} and SMART\textsubscript{BERT}~\cite{smart} show performance gains in comparison to BERT~\cite{Bert} and RoBERTa~\cite{liu2019roberta} across a variety of NLP tasks. XLNet~\cite{xlnet} improves over BERT using auto-regressive pretraining, enabling leaning bidirectional contexts using the expected likelihood over all permutations of the factorization order, and solving BERT's limitations due to neglecting dependency between masked positions. 

The methods and techniques which are BERT-based and proposed in the last couple of years do not incorporate in any way the nature of representations learned and generated by BERT, While for a task like classification, the linguistic features fed to a classifier can be crucial, as they encode relevant information about the input in the embedding space.\newline\indent
Convolutional Neural Networks have proven to be efficient feature extractors which serve well for text classification. Kim Yoon~\cite{kim2014convolutional} suggested training a CNN architecture with filters of multiple sizes, on top of pre-trained Word2Vec~\cite{word2vec} embeddings. The feature maps from the different filters, are concatenated after a max-over-time pooling operation, and passed to a Softmax layer. Zhang et al.~\cite{zhang-wallace-2017-sensitivity} expanded on Kim Yoon's~\cite{kim2014convolutional} model, with an empirical identification of hyper-parameter settings. Kim's CNN for text~\cite{kim2014convolutional} motivates a portion of our work, and inspires a critical part of the proposed method in this paper. DPCNN~\cite{dpcnn} employs deepening of word level CNNs to capture global representations of text, and proposes a deep pyramid convolutional neural network that achieves the best accuracy by increasing the network depth, without increasing computational cost. \newline\indent
Standalone hybrid frameworks which combine attention and CNNs were proposed, in an attempt to leverage their capabilities simultaneously. In HAHNN~\cite{HAHNN}, both word and sentence-level context vectors are generated hierarchically using separate attention blocks, in order to construct a document representation based on the interaction between words and sentences, setting a high-level context vector which serves as representation for a document. A variation of the HAHNN~\cite{HAHNN} method employs a simple convolutional layer after the embedding layer for the purpose of effective feature extraction before the attention blocks.

\section{Model}\label{model}

\begin{figure*}[ht]
    \center
    \hspace*{0.2cm}  
    \begin{tikzpicture}

\draw[red, thick] (0,0.1) rectangle (1.5,0.5);
\draw[red, thick] (0,-0.5) rectangle (1.5,-0.1);
\draw[red, thick] (0,-1.1) rectangle (1.5,-0.7);
\draw[red, thick] (0,-1.7) rectangle (1.5,-1.3);

\draw[blue, thick] (0,-2.3) rectangle (1.5,-1.9);
\draw[blue, thick] (0,-2.9) rectangle (1.5,-2.5);
\draw[blue, thick] (0,-3.5) rectangle (1.5,-3.1);
\draw[blue, thick] (0,-4.1) rectangle (1.5,-3.7);

\draw[green, thick] (0,-4.7) rectangle (1.5,-4.3);
\draw[green, thick] (0,-5.3) rectangle (1.5,-4.9);
\draw[green, thick] (0,-5.9) rectangle (1.5,-5.5);
\draw[green, thick] (0,-6.5) rectangle (1.5,-6.1);

\node[draw,white] at (0.8,0.9) {\color{black}$X \in \mathbb{R}^{12\times768}$};

\draw[black, thin] (-0.2,-6.7) rectangle (1.7,0.6);

\fill[red, thick] (2.5,-0.3) rectangle (4,0.1);
\fill[blue, thick] (2.5,-0.9) rectangle (4,-0.5);
\fill[green, thick] (2.5,-1.5) rectangle (4,-1.1);

\fill[red, thick] (2.5,-2.7) rectangle (4,-2.3);
\fill[blue, thick] (2.5,-3.3) rectangle (4,-2.9);
\fill[green, thick] (2.5,-3.9) rectangle (4,-3.5);

\fill[red, thick] (2.5,-5.1) rectangle (4,-4.7);
\fill[blue, thick] (2.5,-5.7) rectangle (4,-5.3);
\fill[green, thick] (2.5,-6.3) rectangle (4,-5.9);

\draw[red,very thick, dotted] (2.5, 0.1) -- (1.5, 0.5);
\draw[red,very thick, dotted] (2.5,-0.3) -- (1.5,-1.7);
\draw[blue, very thick, dotted] (2.5,-0.9) -- (1.5,-4.1);
\draw[blue,very thick, dotted] (2.5,-0.5) -- (1.5,-1.9);
\draw[green, very thick, dotted] (2.5,-1.5) -- (1.5, -6.5);
\draw[green, very thick, dotted] (2.5, -1.1) -- (1.5,-4.3);

\draw[red,very thick, dotted] (2.5, -2.3) -- (1.5, 0.5);
\draw[red,very thick, dotted] (2.5,-2.7) -- (1.5,-1.7);
\draw[blue, very thick, dotted] (2.5,-3.3) -- (1.5,-4.1);
\draw[blue,very thick, dotted] (2.5,-2.9) -- (1.5,-1.9);
\draw[green, very thick, dotted] (2.5,-3.5) -- (1.5,-4.3);
\draw[green, very thick, dotted] (2.5, -3.9) -- (1.5, -6.5);

\draw[red,very thick, dotted] (2.5, -4.7) -- (1.5, 0.5);
\draw[red,very thick, dotted] (2.5,-5.1) -- (1.5,-1.7);
\draw[blue, very thick, dotted] (2.5,-5.3) -- (1.5,-1.9);
\draw[blue,very thick, dotted] (2.5,-5.7) -- (1.5,-4.1);
\draw[green, very thick, dotted] (2.5,-5.9) -- (1.5,-4.3);
\draw[green, very thick, dotted] (2.5, -6.3) -- (1.5, -6.5);

\draw [red, thick] (4.5, -0.3) rectangle (5.4, 0.25) node [pos=.5]{\color{black}$h^{Q}$};
\draw [blue, thick] (4.5, -0.95) rectangle (5.4, -0.44) node [pos=.5]{\color{black}$s^{Q}$};
\draw [green, thick] (4.5, -1.6) rectangle (5.4, -1.1) node [pos=.5]{\color{black}$v^{Q}$};

\draw [red, thick] (4.5, -2.7) rectangle (5.4, -2.15) node [pos=.5]{\color{black}$h^{K}$};
\draw [blue, thick] (4.5, -3.35) rectangle (5.4, -2.84) node [pos=.5]{\color{black}$s^{K}$};
\draw [green, thick] (4.5, -4) rectangle (5.4, -3.5) node [pos=.5]{\color{black}$v^{K}$};

\draw [red, thick] (4.5, -5.1) rectangle (5.4, -4.55) node [pos=.5]{\color{black}$h^{V}$};
\draw [blue, thick] (4.5, -5.75) rectangle (5.4, -5.25) node [pos=.5]{\color{black}$s^{V}$};
\draw [green, thick] (4.5, -6.38) rectangle (5.4, -5.9) node [pos=.5]{\color{black}$v^{V}$};

\draw[red,very thick, dotted] (4, -0.1) -- (4.5, -0.1);
\draw[blue,very thick, dotted] (4, -0.7) -- (4.5, -0.7);
\draw[green,very thick, dotted] (4, -1.3) -- (4.5, -1.3);

\draw[red,very thick, dotted] (4, -2.5) -- (4.5, -2.5);
\draw[blue,very thick, dotted] (4, -3.1) -- (4.5, -3.1);
\draw[green,very thick, dotted] (4, -3.7) -- (4.5, -3.7);

\draw[red,very thick, dotted] (4, -4.9) -- (4.5, -4.9);
\draw[blue,very thick, dotted] (4, -5.5) -- (4.5, -5.5);
\draw[green,very thick, dotted] (4, -6.1) -- (4.5, -6.1);

\draw [red, thick] (6, -0.3) rectangle (6.8, 0.25) node [pos=.5]{\color{black}$h_{p}^{Q}$};
\draw [blue, thick] (6, -0.95) rectangle (6.8, -0.44) node [pos=.5]{\color{black}$s_{p}^{Q}$};
\draw [green, thick] (6, -1.6) rectangle (6.8, -1.1) node [pos=.5]{\color{black}$v_{p}^{Q}$};

\draw [red, thick] (6, -2.7) rectangle (6.8, -2.15) node [pos=.5]{\color{black}$h_{p}^{K}$};
\draw [blue, thick] (6,-3.35) rectangle (6.8,-2.84) node [pos=.5]{\color{black}$s_{p}^{K}$};
\draw [green, thick] (6, -4) rectangle (6.8, -3.5) node [pos=.5]{\color{black}$v_{p}^{K}$};

\draw [red, thick] (6, -5.1) rectangle (6.8, -4.55) node [pos=.5]{\color{black}$h_{p}^{V}$};
\draw [blue, thick] (6, -5.75) rectangle (6.8, -5.25) node [pos=.5]{\color{black}$s_{p}^{V}$};
\draw [green, thick] (6, -6.38) rectangle (6.8, -5.9) node [pos=.5]{\color{black}$v_{p}^{V}$};

\draw[red,very thick, dotted] (5.4, -0.1) -- (6, -0.1);
\draw[blue,very thick, dotted] (5.4, -0.7) -- (6, -0.7);
\draw[green,very thick, dotted] (5.4, -1.3) -- (6, -1.3);

\draw[red,very thick, dotted] (5.4, -2.5) -- (6, -2.5);
\draw[blue,very thick, dotted] (5.4, -3.1) -- (6, -3.1);
\draw[green,very thick, dotted] (5.4, -3.7) -- (6, -3.7);

\draw[red,very thick, dotted] (5.4, -4.9) -- (6, -4.9);
\draw[blue,very thick, dotted] (5.4, -5.5) -- (6, -5.5);
\draw[green,very thick, dotted] (5.4, -6.1) -- (6, -6.1);

\draw [decorate,
    decoration = {brace, amplitude=4.5pt,raise=1.5pt}] (5.2, 0.31) --  (6.2, 0.31) node [text width = 2cm,above = 4.3pt, midway, align=center, anchor=south,xshift=-0.3ex, black]{Adaptive MaxPooling};
    
\draw [decorate,
    decoration = {brace, amplitude=4.5pt,raise=1.5pt}] (2.4, 0.2) --  (4.05, 0.2) node [text width = 2cm,above = 8pt, midway, align=center, anchor=south,xshift=-0.25ex,yshift=-0.5, black]{Convolutional Filters};
    
\node[draw,white,scale=1.5] at (7.5,-0.7) {\color{black}$\oplus$};
\node[draw,white,scale=1.5] at (7.5,-3.1) {\color{black}$\oplus$};
\node[draw,white,scale=1.5] at (7.5,-5.5) {\color{black}$\oplus$};

\draw[red,very thick, dotted] (6.8, -0.1) -- (7.3, -0.7);
\draw[blue,very thick, dotted] (6.8, -0.7) -- (7.3, -0.7);
\draw[green,very thick, dotted] (6.8, -1.3) -- (7.3, -0.7);

\draw[red,very thick, dotted] (6.8, -2.51) -- (7.3, -3.11);
\draw[blue,very thick, dotted] (6.8, -3.11) -- (7.3, -3.11);
\draw[green,very thick, dotted] (6.8, -3.73) -- (7.3, -3.11);

\draw[red,very thick, dotted] (6.8, -4.92) -- (7.3, -5.52);
\draw[blue,very thick, dotted] (6.8, -5.52) -- (7.3, -5.52);
\draw[green,very thick, dotted] (6.8, -6.12) -- (7.3, -5.52);

\draw[black, thin] (8, -0.3) rectangle (9, -1.2) node [pos=.5]{\color{black}$Q_{conv}$};
\draw[black,very thick, dotted] (7.63, -0.7) -- (8, -0.7);

\draw[black, thin] (8, -2.71) rectangle (9, -3.61) node [pos=.5]{\color{black}$K_{conv}$};
\draw[black,very thick, dotted] (7.63, -3.11) -- (8, -3.11);

\draw[black, thin] (8, -5.12) rectangle (9, -6.02) node [pos=.5]{\color{black}$V_{conv}$};
\draw[black,very thick, dotted] (7.63, -5.52) -- (8, -5.52);

\definecolor{apricot}{rgb}{0.98, 0.81, 0.69}
\definecolor{corn}{rgb}{0.98, 0.93, 0.36}
\definecolor{cyan(process)}{rgb}{0.0, 0.72, 0.92}

\fill[apricot, thick] (9.5, -5.12) rectangle (10.2, -1.2) node [pos=.5, rotate=270]{\color{black}$MultiHeadAtt$};

\draw[black,very thick, dotted] (9, -0.7) -- (9.5, -2.7);
\draw[black,very thick, dotted] (9, -3.11) -- (9.5, -3.11);
\draw[black,very thick, dotted] (9, -5.52) -- (9.5, -3.50);

\draw[black,very thick, dotted] (10.2, -3.11) -- (10.6, -3.11);
\fill[corn, thick] (10.6, -5.12) rectangle (11.3, -1.2) node [pos=.5,rotate=270]{\color{black}$Add\&Norm$};

\draw[black,very thick, dotted] (0.8, 1.1) -- (0.8,1.5);
\draw[black,very thick, dotted] (0.8, 1.5) -- (11,1.5);
\draw[black,very thick, dotted] (11, 1.5) -- (11, -1.2);

\fill[cyan, thick] (11.9, -5.12) rectangle (12.6, -1.2) node [pos=.5,rotate=270]{\color{black}$FF$ $Layer$};
\draw[black,very thick, dotted] (11.3, -3.11) -- (11.9, -3.11);
\draw[black,very thick, dotted] (11.6, -3.11) -- (11.6, -0.6);
\draw[black, very thick, dotted] (11.6,-0.6) -- (13.25,-0.6);
\draw[black,very thick, dotted] (13.25,-0.6) -- (13.25, -1.2);
\fill[corn, thick] (12.9, -5.12) rectangle (13.6, -1.2) node [pos=.5,rotate=270]{\color{black}$Add\&Norm$};
\draw[black,very thick, dotted] (12.6, -3.11) -- (12.9, -3.11);
\draw[black,very thick, dotted] (13.6, -3.11) -- (14.4, -3.11);
\node[draw,white] at (13.95,-2.8) {\color{black}$X_{0}$};
\draw[black, very thick, dotted] (14.3,-2.5) -- (14.3,-3.75);
\draw[black,very thick, dotted] (14.3,-3.11) -- (14.6,-3.11);
\draw[black,very thick, dotted] (14.3,-3.75) -- (14.6,-3.75);
\draw[black,very thick, dotted] (14.3,-2.5) -- (14.6,-2.5);

\fill[apricot, thick] (14.6, -5.12) rectangle (15.3, -1.2) node [pos=.5, rotate=270]{\color{black}$MultiHeadAtt$};
\draw[black, very thick, dotted] (15.3, -3.11) -- (15.7,-3.11);
\fill[corn, thick] (15.7,-5.12) rectangle (16.3, -1.2) node [pos=.5,rotate=270]{\color{black}$Norm$ $Layer$};
\draw[black, very thick, dotted] (16.3,-3.11) -- (16.8,-3.11);
\draw[black, very thick, dotted] (16.55,-3.11) -- (16.55, -0.63);
\fill[cyan, thick] (16.8, -5.12) rectangle (17.4, -1.2) node [pos=.5,rotate=270]{\color{black}$FF$ $Layer$};
\draw[black, very thick, dotted] (17.4,-3.11) -- (17.8,-3.11);
\fill[corn, thick] (17.8,-5.12) rectangle (18.4, -1.2) node [pos=.5,rotate=270]{\color{black}$Add\&$Norm};
\draw[black, very thick, dotted] (16.55, -0.63) -- (18.1,-0.63);
\draw[black, very thick, dotted] (18.1, -0.63) -- (18.1, -1.2);
\draw[black, very thick, dotted] (18.4,-3.11) -- (18.7, -3.11);
\node[draw, white,rotate=270] at (18.9,-3.11){\color{black}$Output$};

\draw [decorate,
    decoration = {brace, amplitude=4.5pt,raise=1pt}] (13.7, -5.4) --  (9.4, -5.4) node [pos=0.5,below = 4pt, black]{$First$ $Trans\-Enc$ $layer$};
    
\draw [decorate,
    decoration = {brace, amplitude=4.5pt,raise=1pt}] (18.7, -5.4) --  (14.2, -5.4) node [pos=0.5,below = 4pt, black]{$Second$ $Trans\-Enc$ $layer$};

\draw [decorate,
    decoration = {brace, amplitude=4.5pt,raise=1pt}] (7.75, -6.6) --  (2.3, -6.6) node [pos=0.5,below = 4pt, black]{3 Replicas of $CNN[CLS]$};

\end{tikzpicture}
    \caption{
    \centering
    Model architecture : CNN-Trans-Enc obtains Queries, Keys, and Values matrices from convolution operations applied on the input $X$, which represents 12 stacked BERT $[CLS]$ representations. The $CNN[CLS]$ submodule is replicated three times. We apply an Adaptive-Max Pooling and concatenation ($\oplus$) on the resulting feature maps to obtain $Q_{conv}, K_{conv}, V_{conv} \in \mathbb{R}^{12 \times d_{m}}$ where $d_m$ is the embedding dimension. $Q_{conv}$, $K_{conv}$, and $V_{conv}$ are then directly fed to a stack of two Transformer-Encoder~\cite{attention} layers.}
    \label{fig:1}
\end{figure*}
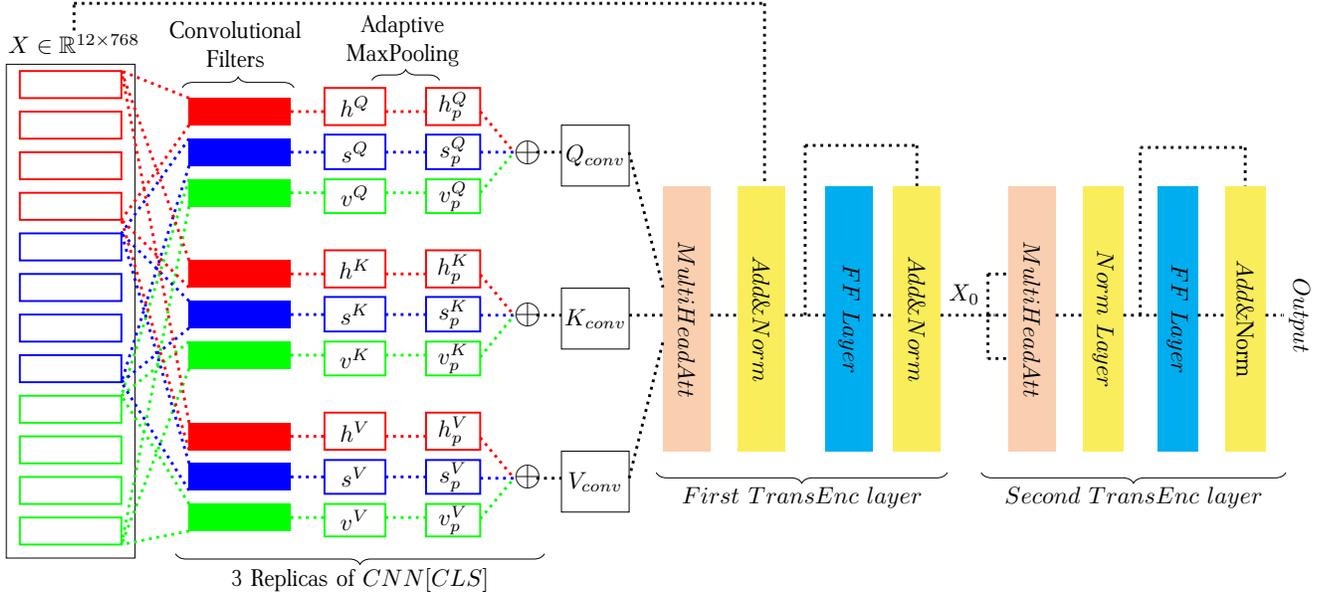

The model architecture, shown in Fig~\ref{fig:1}, consists of a representation layer, where BERT~\cite{Bert} is used to obtain fixed $[CLS]$ representations from its 12 layers. In our experiment, the pre-trained BERT-base model is exclusively employed as an embedding layer, and $[CLS]$ representations from all layers are concatenated to form a tensor $X \in R^{12 \times 768}$ for each text sample in a dataset $D$. This work presents a more practical downstream model for classification tasks, added on top of BERT. CNN-Trans-Enc is a variant of the original Transformer-Encoder~\cite{attention}, which uses convolution based submodules to obtain Q, K, and V matrices instead of linear projections of the input to the embedding space. To obtain more efficient and meaningful representations which contain linguistic features learned by all BERT layers, we treat the 12 $[CLS]$ representations for a sample $s$, as $12$ channels. We use three convolutional filters set to a length of $l$, $W_{1}, W_{2}, W_{3} \in \mathbb{R}^{4 \times l}$. Each filter takes in every $4$ adjacent channels, and each feature $h_{i}$, $s_{i}$, or $v_{i}$ could be computed as follows:
\begin{equation}\label{eq:CNN[CLS]}
\centering
\begin{aligned}
h_{i} = f(W^{T}_{1} \cdot X_{i:i+l-1} + B_{h})\\
s_{i} = f(W^{T}_{2} \cdot X_{i:i+l-1} + B_{s})\\
v_{i} = f(W^{T}_{3} \cdot X_{i:i+l-1} + B_{v})
\end{aligned}
\end{equation}
Where $B$ is the bias term, $f$ is the  hyperbolic tangent non-linear function, $i$ is the position of the current convolution operation increase by a stride of 2, and $X_{i:i+l-1} \in \mathbb{R}^{4 \times l}$ is a window of length $l$ from convolution position $i$ to position $i+l-1$. Three feature maps $h$, $s$, and $v$ are then formed with:
 \begin{equation}\label{eq:cnn12concat}
     \centering
     \begin{aligned}
        h = [h_{0}, h_{1},...., h_{n}]\\
        s = [s_{0}, s_{1},...., s_{n}]\\
        v = [v_{0}, v_{1},...., v_{n}]
     \end{aligned}
 \end{equation}
Where $n = \frac{768 - kernel\_size}{2} + 1$. Adaptive-Max pooling is used to reduce $h,s,v \in \mathbb{R}^{4\times 382}$ to $h_{p},s_{p},v_{p}\in \mathbb{R}^{4 \times d_{m}}$. The resulting pooled feature-maps $h_{p}, s_{p}$ and $v_{p}$ are then concatenated over the first dimension to form the desired $Q_{conv}$, $K_{conv}$, or $V_{conv}$ matrix $[h_{p}; s_{p}; v_{p}]$. We refer to this CNN-based module as $CNN[CLS]$. We employ three replicas of $CNN[CLS]$ to generate $Q_{conv}, K_{conv}, V_{conv} \in \mathbb{R}^{12\times d_{m}}$ simultaneously in parallel, where $d_{m}$ is the chosen embedding dimension~\cite{attention}.
After generating $Q_{conv}, K_{conv}$, and $V_{conv}$, we feed them into a first Transformer-Encoder~\cite{attention}, starting with a multi-head attention:
\begin{equation}\label{eq:mha}
    \begin{split}
    &\operatorname{Multi-head}(Q_{conv},K_{conv},V_{conv}) = \\
    &Concat(head_1, ... , head_h)W^O
    \end{split}
\end{equation}
\begin{equation}
\hspace{-2mm}
    head_k = Attention(Q_{conv}{W_k}^Q, K_{conv}{W_k}^K, V_{conv}{W_k}^V)
\end{equation}
${W_k}^{Q}$, ${W_k}^{K}$, ${W_{k}}^{V} \in \mathbb{R}^{d_m \times d_v}$ and $W^{O} \in \mathbb{R}^{d_{m} \times 768}$ are parameter matrices and 
$Attention$ is defined as:
\begin{equation}
    \begin{split}
        Attention(Q,K,V) = softmax(\frac{QK^{T}}{\sqrt{d_{k}}})V
    \end{split}
\end{equation}
 \newline\indent
The main objective behind using a Transformer-Encoder~\cite{attention} that takes in $[CLS]$ representations from all layers, is to capture linguistic features learned by all layers. And to integrate the findings reported by Jawahar et al.~\cite{jawahar-etal-2019-bert}, the Scaled-Dot Product attention is used in this scope to capture similarities between $[CLS]$ representations from all layers. Intuitively, using two Transformer-Encoder layers, and reducing the size of the output projection matrix for the second one, results in a representation that encodes multiple linguistic features captured by BERT layers, in one fairly small vector. We safely remove the positional encoding since the Transformer-Encoder receives the same order of $[CLS]$ representations from the exact same BERT layers which are frozen and exclusively used as a feature extractor. The output of the first Transformer-Encoder block is denoted as $X_{0} \in \mathbb{R}^{12\times 768}$, and it is of the same size as the input $X$. \newline\indent 
The Second Transformer-Encoder layer is almost similar to the first one, they only differ in the Multi-Head Attention operation. The projection matrix $W_{outdim}^O \in \mathbb{R}^{(12d_m)\times outdim}$ is applied to the vectorized output of the $\operatorname{Multi-head}$ attention, where $outdim$ is the representation dimension we obtain from the second Transformer-Encoder layer. We remove the residual connection linking the input to the Multi-head Attention in the second layer and its output, due to the incompatible shapes caused by the projection matrix $W_{outdim}^O$.
The Multi-Head Attention equation is therefore reformulated as:
\begin{equation}\label{eq:mhacnn}
    \operatorname{Multi-head}(Q_{2},K_{2},V_{2}) = ZW_{outdim}^O
\end{equation}
\begin{equation}\label{vectorizatoin}
    Z = \mathbf{vec}(Concat(head_1, ... , head_h))
\end{equation}
\begin{equation}
    head_k = Attention(Q_{2}{W_k}^Q, K_{2}{W_k}^K, V_{2}{W_k}^V)
\end{equation}
\noindent
Where $Q_{2},K_{2}$, $V_{2} \in \mathbb{R}^{12
 \times d_{m}}$ are obtained with linear projections of the output of the first Transformer-Encoder layer $X_{0} \in \mathbb{R}^{12\times 768}$, through projection matrices $W_{Q}, W_{K}, W_{V} \in \mathbb{R}^{768\times d_{m}}$. $\mathbf{vec}$ is the vectorization of $Concat(head_1, ... , head_h) \in \mathbb{R}^{12\times d_{m}}$ to a vector $Z \in \mathbb{R}^{12d_{m}}$. 
\comment{
\begin{table}[htbp]
\begin{center}
\caption{Post-Layer Normalization Transformer-Encoder proposed in~\cite{attention}, $Q_2,K_2,V_2 \in \mathbb{R}^{12 \times d_{m}}$: Obtained from linear projections of output of the first Transformer-Encoder layer where $d_m$ is the embedding dimension. $MultiHeadAtt$: Multi-Head Attention~\cite{attention}, 
$LayerNorm$: Layer Normalization~\cite{layernorm}, addition in lines 4 is a residual connection introduced in~\cite{residual}}
\label{trans-enc2-table}
\begin{tabular}{p{1cm} p{6cm}}
\toprule
\textbf  & {Second Transformer-Encoder layer} \\
\midrule
 1 & $X_1 = MultiHeadAtt(Q_2,K_2,V_2)$ \\\
 2 & $X_2 = LayerNorm(X_1)$ \\\
 3 & $X_3 = Tanh(X_2)$ \\\
 4 & $X_{4} = X_2 + X_3$ \\\
 5 & $Output = LayerNorm(X_{4})$ \\
\bottomrule
\end{tabular}
\end{center}
\vspace{-4mm}
\end{table} 
}
$Output \in \mathbb{R}^{outdim}$ is a vector representation of the whole concatenation of the $[CLS]$ representations obtained from BERT, which encodes linguistic features learned by all BERT layers, chosen and pooled using the $CNN[CLS]$ modules, and reduced to a small vector by exploiting the attention mechanism in the Transformer-Encoder, which allows for similarity measurement, and dimensionality reduction by altering the output projection matrix of the second Transformer-Encoder layer. Using three separate CNN-based modules on the output of BERT~\cite{Bert} allows for more effective feature extraction, taking in consideration the features which are considered to be most relevant by the convolution and pooling operations. The attention mechanism is then applied to the resulting feature maps to generate a vector comprising different kinds of linguistic features learned by BERT, and serves as a fairly small representation for classification tasks which helps achieve competitive results without fine-tuning BERT.
The last layer is a Softmax function which receives $Output$ and generates probability distributions over classes according to:
\begin{equation}\label{eq:Softmax2}
    p(c|Output) = Softmax(\mathbf{W} \cdot Output)
\end{equation}
Where $\mathbf{W}$ is a parameter matrix, and $Output \in \mathbb{R}^{outdim}$ is the representation obtained from applying CNN-Trans-Enc on the 12 $[CLS]$ representations from all layers of BERT.

\section{Datasets and Experimental Setup}

Survey~\cite{holistic} is a holistic overview of $Text$ and $Sentiment$ $Classification$ methods and processes. We follow the scheme for achieving state-of-the-art $Document$ $Classification$ as follows:

\begin{itemize}
    \item  Datasets are chosen according to current state-of-the-art in sentiment and topic classification, and news categorization. We used six supervised datasets both for binary and multi-class classification, section~\ref{datasets} and TABLE~\ref{dataset-tab}.
    \item Evaluation is conducted using K-fold Cross Validation using random splits, and repeated over five random seeds, average Cross-Validation results are reported in TABLE~\ref{results}. Alongside our proposed architecture, we evaluate other four models on top of fixed representations from BERT, according to different design choices, and use ASO~\cite{aso} test for statistical Significance Testing between pairs of models in our experiment.
\end{itemize}

\subsection{Datasets}\label{datasets}
We evaluate our work on multiple document classification datasets, including four sentiment analysis datasets, one news classification dataset and one topic classification dataset, shown in TABLE~\ref{dataset-tab}: 
\begin{table}[htbp]
\begin{center}
\caption{Dataset statistics, C: Number of classes, T: Task, N: Number of samples}
\label{dataset-tab}
\begin{tabular}{p{2cm} p{1.2cm} p{1.2cm} p{1.2cm}}
\toprule
\textbf{Dataset}&{C}&{T}&{N} \\
\midrule
IMDB& 2 & SA & 50k \\
Amazon& 2 & SA & 1M \\
YELP& 5 & SA & 1M \\
SST-5& 5 & SA & 11861 \\
AG& 4 & NC & 127600 \\
DBPedia& 14 & TC & 630,000 \\
\bottomrule
\end{tabular}
\end{center}
\vspace{-4mm}
\end{table}

\begin{itemize}
    \item IMDB~\cite{imdb}: IMDB movie reviews dataset, is a collection of 50,000 reviews, with a maximum of 30 reviews per movie. A review is negative if it has a score $<$ 4 out of 10, and a positive review has score $>$ 7 out of 10. 
    \item Yelp-Reviews~\cite{charcnn}: This is a dataset for business reviews and ratings, currently containing 8,635,403 samples with 5 labels corresponding to the scores associated with the sentiments expressed in the reviews.
    We use a subset of 1 million reviews from the Yelp-Reviews dataset and we make sure we have a balanced set, with 200 thousand reviews for each label.
    \item SST-5~\cite{sst-5}: Stanford Sentiment Treebank is a Sentiment Scoring corpus, which contains parse trees for all samples, We use 11861 samples, with labels binned in 5 bins from 0 to 4.
    
    \item Amazon-Polarity~\cite{charcnn}: The Amazon Polarity dataset, contains product reviews from Amazon, up to march 2013 it contained up to 35  million reviews with binary labels representing the corresponding sentiments as positive or negative classes. We use 1 million samples with a balanced distribution of 500 thousand samples for each label.
    
    \item AG's news corpus~\footnote{\url{http://groups.di.unipi.it/~gulli/AG_corpus_of_news_articles.html}}: AG is a corpus of news articles containing 127,600 samples, distributed over 4 classes, obtained from more than 2,000 news sources by ComeToMyHead, an academic news search engine. We use the entire number of samples for k-fold cross validation in our experiment.
    
    \item DBPedia 14~\cite{DBpedia14}: DBPedia is formed by extracting structured information from Wikipedia, it contains 630,000 samples, and 14 ontology classes. We use the entirety under k-fold cross validation in our experiment.
\end{itemize}

\subsection{Training and Hyperparameters}\label{title:training}
 We use the BERT-base model~\cite{Bert}, which is pre-trained and available online by the HuggingFace team\footnote{\url{https://huggingface.co/bert-base-uncased}}, with a hidden size of 768, 12 layers, and 12 attention heads. We use the pre-trained BERT model once for each dataset and generate $[CLS]$ representations from all 12 layers, for all samples in our datasets. This means that if a dataset has a number of samples $N$, this is results in dataset $D\in \mathbb{R}^{N\times 12 \times 768}$, where the second dimension $12$, represents $12$ $[CLS]$ representations from $12$ layers.\newline\indent
Our proposed architecture, which is trained on top of fixed $[CLS]$ representations from BERT~\cite{Bert}, combines convolution-based modules, and two stacked Transformer-Encoder layers. In this work,we employ $h = 20$ parallel attention heads for both Trans-Enc layers and we set $d_{k} = d_{v} = d_{m} / h = 19$. The embedding dimension $d_{m}$ is set to $380$ and $outdim$ to $320$.\newline\indent
The $CNN[CLS]$ convolutional modules used to obtain $Q_{conv}$, $K_{conv}$, and $V_{conv}$, are trained with hyperparameter choices that are heavily influenced by the the work conducted by Kim Yoon~\cite{kim2014convolutional} and Zhang et al.~\cite{zhang-wallace-2017-sensitivity}. We give the baseline CNN configuration described in TABLE~\ref{cnn-configs}.

\begin{center}
\begin{table}[H]
\begin{center}
\caption{Baseline configuration. Adam: Adam Optimizer~\cite{Adam}, filter-sizes: both architectures have three convolutional filters. Xavier-Init: Xavier-Initialization~\cite{xavier}, Tanh:  hyperbolic tangent non-linear function.}
\label{cnn-configs}
\begin{tabular}{p{3.5cm}  p{1.5cm}}
\toprule
\textbf Description & Values \\
\midrule
 optimizer & Adam \\\
 learning-rate & 0.0001 \\\
 filter-length $l$ & 5  \\\
 input-channels & 12  \\\
 initialization & Xavier-Init \\\
 activation function& Tanh\\
 Dropout ratio& 0.3\\
\bottomrule
\end{tabular}
\end{center}
\vspace{-1mm}
\end{table}
\end{center}

 The Post-LN Transformer-Encoder~\cite{attention}~\cite{postln} from which our two Trans-Enc layers are inspired, suffers from instability when optimized with large learning rates and requires careful optimization choices. For an accurate evaluation, we follow the training paradigm used with the original Transformer architecture~\cite{attention}, and emphasized by Xiong et al.~\cite{postln}.
For both Transformer-Encoder layers, we use Adam optimizer~\cite{Adam}, with $\beta_1 = 0.9$, $\beta_2 = 0.98$, and $\epsilon = 10^{-9}$.
The learning rate is increased during a warmup-stage according to:
\begin{equation}
    lr(t) = \frac{t}{T_{warmup}}lr_{max}, t\leq T_{warmup}
\end{equation}

We set $T_{warmup} = 1000$, $lr_{max} = 0.001$, and adapt the batch-size for all datasets to have $6000$ training steps for each one. After $T_{warmup}$, inverse square-root decay is used as a scheduler to set the learning rate. Xavier initialization~\cite{xavier} is consistently used throughout our experiment.\newline\indent
 We replicate the experiment five times using different random seeds. Each replication constitutes five runs of a 5-fold Cross-Validation. We report averages of Cross-Validation accuracy means over the five seeds in TABLE~\ref{cross-val-table}.
\subsection{Model Variations}

\begin{table*}[ht]
\begin{center}
 \caption{Results of our CNN-Trans-Enc model against other methods. Results recorded on the first five rows concern the models implemented by the authors and tested according to the methodology described in this paper (recorded 5-fold Cross-Validation accuracies as described in \ref{title:training}). $*$ means results are reported on their papers. $-$ means not reported. Results without $*$ are obtained from our experiments. RoBERTa-base~\cite{liu2019roberta} and ALBERT~\cite{albert} have no reported performance on the used datasets, and their performance on the IMDB dataset is obtained from our experiment, due to the small size of IMDB and difficulty to train both of them.}
\label{cross-val-table}
\begin{tabular}{p{4.5cm} | p{1.2cm} p{1.2cm} p{1.2cm} p{1.2cm} p{1cm} p{1.2cm}}
\toprule
\textbf {Model} & IMDB & YELP-5 & SST-5 & AMZ-2 & AG & DBP \\
\midrule
BERT-last+Softmax~\cite{howtofinetunbert} & 93.21 & 70.1 & 49.15 & 93.2 & 92.05 & 99.10 \\\
BERT-last+Kim-CNN~\cite{kim2014convolutional} & 94.3 & 72.05 & 51.23 & 96.45 & 93.21 & 99.28\\\
BERT-12+CNN[CLS] & 88.0 & 65.3 & 48.1 & 91.2 & 86.98 & 98.73 \\\
BERT-12+Trans-Enc~\cite{attention} & 94.6 & 75.33 & 53.01 & 96.9 & 93.40 & 99.43 \\\
BERT-12+CNN-Trans-Enc(Ours) & 95.7 & \textbf{82.23} & 56.03 & \textbf{98.05} & 95.49 & \textbf{99.51}\\
\midrule
BERT-ITPT-FiT~\cite{howtofinetunbert} & 95.63* & 70.58* & - & - & - & - \\\
DistilBERT~\cite{Sanh2019DistilBERTAD} & 92.82* & - & - & - & - & - \\\
RoBERTa-base~\cite{liu2019roberta} & 93.78 & - & - & - & - & - \\\
ALBERT~\cite{albert} & 88.69 & - & - & - & - & - \\\
HAHNN(CNN)~\cite{HAHNN} & 92.26* & 73.28* & - & - & - & - \\\
HAHNN (TCN)~\cite{HAHNN} & 95.17* & 72.63* & - & - & - & - \\\
XLNet~\cite{xlnet} & 96.8* & 72.95* & - & 97.89* & \textbf{95.55}* & 99.40* \\\
BERT-base~\cite{munikar2019finegrained, lin-etal-2021-bertgcn} & 95.63* & 70.80* & 53.2* & - & 95.20* & 99.35*\\\
BERT-large~\cite{munikar2019finegrained, Xie2020UnsupervisedDA, howtofinetunbert} & 95.79* & 70.68* & 55.5* & 97.37* & 95.34* & 99.39*\\\
BERT-large-finetune-UDA~\cite{Xie2020UnsupervisedDA} & 95.8* & 67.92* & - & 96.5* & - & - \\\
RoBERTa-large+Self-Explaining~\cite{roberta+selfexp} & - & - & \textbf{59.1}* & - & - & -\\\
DPCNN~\cite{dpcnn} & - & 69.42* & - & 96.68* & 93.13* & 99.12*\\\
CCCapsNet~\cite{ren2018compositional} & - & - & - & 94.96* & 92.39* & 98.72* \\\
EXAM~\cite{exam} & - & - & - & 95.5* & - & - \\\
DRNN~\cite{wang-2018-disconnected} & - & - & - & 96.49* & 94.47* & 99.19 \\
NB-weighted-BON + dv-cosine~\cite{NB} & \textbf{97.42}* & - & - & - & - & - \\\
EFL~\cite{EFL} & 96.1* & - & - & - & 86.10* & - \\\
GraphStar~\cite{graphstar} & 96.0* & - & - & - & - & - \\\
ULMFiT~\cite{ulmfit} & 95.40* & 70.02* & - & - & 94.99* & 99.20* \\\
VDCNN~\cite{vdcnn} & - & 64.72* & - & 95.72* & 91.33* & 98.71* \\\
ULMFiT (small-data)~\cite{ulmfit-smalldata} & - & 67.60* & - & 96.10* & 93.70* & 99.20* \\\
CharCNN~\cite{charcnn} & - & 62.05* & - & 95.07* & 91.45* & 98.63* \\\
DNC + CUW~\cite{DNC+CUW} & - & - & - & - & 93.90* & 99.00* \\

\bottomrule
\end{tabular}
\newline
\end{center}
\end{table*}
Alongside CNN-Trans-Enc, we experiment with the $CNN[CLS]$, and the two Transformer-Encoder layers (Trans-Enc) separately to understand the effect of each one. They are both trained on top of 12 $[CLS]$ representations from BERT, while freezing BERT layers during training. 

\begin{itemize}
    \item {$\mathbf{CNN[CLS]}$}: implemented as described in \ref{model}, feeding the pooled features from $h,s$, and $v$ feature maps into a softmax classifier, which outputs probabilities for correct class labels. 
    \item \textbf{Trans-Enc}: implemented and trained in the same way as the two Transformer-Encoder layers in \ref{model}, only the $Q$, $K$, and $V$ matrices are obtained using linear projections as proposed by Vaswani et al.~\cite{attention} unlike in CNN-Trans-Enc. $Output$ is directly fed to a softmax classifier as described in equation~\ref{eq:Softmax2}. Training procedure conforms to hyperparameter choices mentioned in \ref{title:training} for both Transformer-Encoder layers. 
\end{itemize}

We also conduct an experiment using the last $[CLS]$ representation from BERT, fed to a Convolutional Neural Network (Kim Yoon~\cite{kim2014convolutional} and Zhang et al.~\cite{zhang-wallace-2017-sensitivity}). A Softmax classifier (equation~\ref{eq:Softmax}) is also trained and evaluated using the last $[CLS]$ representation. the later pair of models is used to evaluate a single $[CLS]$ representation using a CNN, and a simple softmax classifier. This comparison serves as an empirical proof of the feature extraction abilities of CNNs, and how they affect the performance in document classification.

\begin{itemize}
    \item \textbf{Kim-CNN}: Instead of the concatenation of word embeddings $x_1 \oplus x_2 \oplus ... \oplus x_i \oplus ... \oplus x_n$, where $x_i \in \mathbb{R}^{d}$, and $n$ is the sentence length, we feed the last $[CLS]$ representation from BERT to three convolution filters $W_{1} \in \mathbb{R}^{l_{1}}$, $W_{2} \in \mathbb{R}^{l_{2}}$, and $W_{3} \in \mathbb{R}^{l_{3}}$. Let $X \in \mathbb{R}^{768}$ be the last $[CLS]$ sentence representation, the application of the convolutional filters is performed on windows of lengths $l_{1}$, $l_{2}$, and $l_{3}$ according to:
    \begin{equation}
\centering
\begin{aligned}
c_{i} = f(W^{T}_{1} \cdot X_{i : i+l_{1} - 1}  + B_{c})\\
d_{i} = f(W^{T}_{2} \cdot X_{i : i+l_{2} -1} + B_{d})\\
k_{i} = f(W^{T}_{3} \cdot X_{i : i+l_{3} -1} B_{k})
\end{aligned}
\end{equation}
 Where $B$ is the corresponding bias term and $i$ is decided by a sliding operation adding a stride of 2 at each convolution step, and f is the hyperbolic tangent non-linear function. $l_{1}, l_{2}$ and $l_{3}$ are window lengths with values $5$, $10$, and $15$ respectively. The resulting feature maps $c$, $d$, and $k$ are formed with simple concatenations:
 \begin{equation}
     \centering
     \begin{aligned}
        c = [c_{0}, c_{1},...., c_{n}]\\
        d = [d_{0}, d_{1},...., d_{n}]\\
        k = [k_{0}, k_{1},...., k_{n}]
     \end{aligned}
 \end{equation}
 
 Where $n = \frac{768 - kernel\_size}{2} + 1$. Then we apply an Adaptive-Max pooling to capture higher values corresponding to essential features in the resulting feature maps $c$, $d$, and $k$. The pooled features are concatenated, fed to a linear classifier followed by a softmax (equation~\ref{eq:Softmax}). Xavier-Init~\cite{xavier}, Adam~\cite{Adam} with a learning rate of $0.001$, and Hyperbolic Tangent Non-Linear Function are used to train Kim-CNN in this experiment. 
 
 \item \textbf{Softmax}: The last fixed $[CLS]$ representation is directly fed to the softmax equation~\ref{eq:Softmax}, and optimized to predict correct class probabilities.
\vspace{-1mm} 
\end{itemize}

\section{Results and Discussion}\label{results}

 We built on a conjecture, that different linguistic features which are distributed over different layers, makes the $[CLS]$ representation from the last layer only partially representative of the sentence. The Transformer-Encoder~\cite{attention} prominently uses Multi-Head attention to derive an output based on the similarities between the input elements. Also, the work conducted by Kim Yoon~\cite{kim2014convolutional}, and Zhang et al.~\cite{zhang-wallace-2017-sensitivity} proves Convolutional Neural Networks serve as optimal feature extractors for text represented with static embeddings like Word2Vec~\cite{word2vec}. CNN-Trans-Enc harnesses both the feature extraction abilities of convolutional neural networks, and the similarity measurement hallmark of the attention mechanism prominently used in Transformers~\cite{attention}.
 By altering the last projection matrix size $W^O$ of the second Transformer-Encoder layer, we optimize the architecture to learn a representation which encodes the different types of linguistic features learned by different layer regions of BERT in varying aggregations (according to~\cite{jawahar-etal-2019-bert}).\newline\indent
On the IMDB dataset, BERT-12+CNN-Trans-Enc does not show performance gains over XLNet~\cite{xlnet} as it scores exactly the same as BERT-ITPT-FiT~\cite{howtofinetunbert}, achieving $98.23\%$ of the current state-of-the-art performance (NB-weighted-BON + dv-cosine~\cite{NB}). BERT-12 + CNN-Trans-Enc outperforms DistilBERT~\cite{Sanh2019DistilBERTAD}, RoBERTa-base~\cite{liu2019roberta}, and ALBERT~\cite{albert}.\newline\indent
On the YELP-5 dataset, both BERT-12+Trans-Enc and BERT-12+CNN-Trans-Enc outperform BERT-ITPT-Fit~\cite{howtofinetunbert}, XLNet~\cite{xlnet}, and HAHNN(CNN)~\cite{HAHNN}, and BERT+CNN-Trans-Enc achieves new state-of-the-art with $82.23\%$, outperforming the current state-of-the-art HAHNN(CNN)~\cite{HAHNN} by a margin of $\approx$ $8.9\%$.\newline\indent
On the SST-5 dataset, our method outperforms BERT-large~\cite{munikar2019finegrained}~\cite{Xie2020UnsupervisedDA} by $\approx$ $1.0\%$, obtaining $56.03\%$. BERT-12+CNN-Trans-Enc achieves $94.8\%$ of the current state-of-the-art performance on the SST-5 dataset, RoBERTa-large+Self-Explaining~\cite{roberta+selfexp} with $59.1\%$.\newline\indent 
BERT+CNN-Trans-Enc also achieves new state-of-the-art on the Amazon-Polarity dataset, exceeding XLNet~\cite{xlnet} by a margin of $\approx$ $0.16\%$, with a 5-fold cross-validation accuracy of $98.05\%$ on only a 1M sample subset of the dataset.\\ \indent
On the AG news dataset BERT+CNN-Trans-Enc achieves $99.94\%$ of the current state-of-the-art performance (XLNet~\cite{xlnet}), and outperforms it on DBPedia~\cite{DBpedia14} by $\approx$ $0.11\%$ points, achieving new state-of-the-art with a $99.51\%$ accuracy.

\subsection*{\textbf{Effect of CNNs to generate QKV matrices:}}

\begin{table*}[ht]
\begin{center}
\caption{$\epsilon^{min}$ scores between all pairs of architectures used on top of static BERT representations. To avoid confusion, the $-$ is used for cases concerning the same model. The values indicate the comparison between the model names in the leftmost column, with the ones in the upper row. $\epsilon^{min} =0.0$ means that the model on the leftmost column is better than the one on the upper row, and $\epsilon^{min} = 1.0$ means the inverse.}
\label{ASO-table}
\begin{tabular}{p{2.5cm} | p{2cm} p{2cm} p{2cm} p{2cm} p{2cm}}
\toprule
\textbf {Model} & CNN-Trans-Enc & Trans-Enc & Softmax & Kim-CNN & CNN[CLS]\\
\midrule\
CNN-Trans-Enc & - & 0.0 & 0.0 & 0.0 & 0.0 \\\
Trans-Enc & 1.0 & - & 0.0 & 0.0 & 0.0 \\\
Softmax & 1.0 & 1.0 & - & 1.0 & 0.0 \\\
Kim-CNN & 1.0 & 1.0 & 0.0 & - & 0.0 \\\
CNN[CLS] & 1.0 & 1.0 & 1.0 & 1.0 & - \\
\bottomrule
\end{tabular}
\newline
\end{center}
\end{table*}

The results depicted in TABLE~\ref{cross-val-table} show that for the last $[CLS]$ representation, BERT-Feat+Kim-CNN~\cite{kim2014convolutional} outperforms BERT-Feat+Softmax~\cite{howtofinetunbert} on all six datasets. This pairwise comparison suggests that for a $[CLS]$ representation from one layer, which is a vector of length 768, Kim-CNN~\cite{kim2014convolutional} serves as a better feature extractor than a simple Dense Network which outputs logits to a Softmax function. We based our initial assumption, on recorded results from previous work on topics that compare CNN-based architectures to simple approaches such as Dense Networks, and the results depicted in TABLE~\ref{cross-val-table} show that projecting a $[CLS]$ representation from BERT to an embedding space using a matrix of trainable parameters with a predefined shape, is comparatively inadequate. In our model CNN-Trans-Enc, we use replicas of CNN[CLS] separately, to produce $Q_{conv}$, $K_{conv}$, and $V_{conv}$ feature maps which are used equivalently in the same way as in the original Transformer-Encoder~\cite{attention}. As discussed before, CNN[CLS] doesn't perform well on its own, but using its output feature maps as inputs to an Attention-based model, seems to exploit similarities between $[CLS]$ representations from the 12 layers more efficiently. CNN-Trans-Enc outperforms the Trans-Enc~\cite{attention}, and achieves higher accuracies by appreciable differences on all six datasets. \newline\indent
In particular, CNN-Trans-Enc (Our model), exceeds the Softmax classifier by $\approx$ 2 points on the IMDB, $\approx$ 5 on Amazon-Polarity, $\approx$ 12 on the YELP-5 datasets, $\approx$ 7 points on SST-5, $\approx$ 3.24 on AG;s news, and $\approx$ 0.4 on DBPedia-14. The CNN[CLS] blocks, used to produce $Q_{conv}$, $K_{conv}$, and $V_{conv}$, contribute with a significant improvement on all datasets, compared to the Trans-Enc results shown in TABLE~\ref{cross-val-table}. 
Since both models (CNN-Trans-Enc and Trans-Enc) share the same design and hyperparameter settings, and only differ in the projection philosophy of the input to the embedding space, it is safe to say that using a Convolutional Neural Network inside a Transformer-Encoder~\cite{attention}, results in significant performance gains for text classification using $[CLS]$ representations from all layers of BERT~\cite{Bert}.

\subsection*{\textbf{Statistical Significance Testing}}

In order to investigate the statistical significance of our results, we used ASO (Almost Stochastic Order)~\cite{aso}, to compare all pairs of models based on five random seeds, stacking CV average means obtained from empirical evaluation from all datasets. Each comparison uses ASO with a confidence level of $\alpha = 0.05$ (before adjusting for all pair-wise comparisons using the Bonferroni correction). $\epsilon^{min}$ values for all pairs are depicted in TABLE~\ref{ASO-table}. Looking at the Significance Test results, we can see that CNN-Trans-Enc is the model with the overall best performance. We also notice that the Softmax approach only outperforms the standalone CNN[CLS] model. Kim-CNN~\cite{kim2014convolutional}, when used on top of the last $[CLS]$ representations from BERT~\cite{Bert}, improves over the Softmax classifier on large datasets.

\section{Conclusion}
In this paper, we investigate the effect of using $[CLS]$ representations from all layers of BERT (which are kept frozen during training) on the classification performance. We propose a CNN-Enhanced Transformer-Encoder to obtain more generalizable representations through convolutional layers and capture similarities between representations from all BERT layers and compute an average using the Multi-Head Attention mechanism. Our model outperforms the vanilla Transformer-Encoder and other approaches used in text classification on six benchmark datasets. BERT+CNN-Trans-Enc achieves new state-of-the-art results on YELP-5, Amazon-Polarity, and DBPedia-14 datasets exceeding large models which are fine-tuned on the mentioned datasets, using limited hardware resources. We validated our results using a Statistical Significance Test and highlight our findings with empirical analysis of the architectures compared in our experiment. In the future, we aim to explore architectures with more generalization and characterization abilities, both for language and vision, and incorporate hybrid techniques, that combine vision, and language-representation methods, to facilitate and improve results for both domains in multiple tasks.

\bibliography{refs.bib}
\bibliographystyle{plain}
\end{document}